\documentclass[runningheads]{llncs}
\usepackage{graphicx}
\usepackage{multirow}
\usepackage[colorlinks,linkcolor=red, anchorcolor=blue, citecolor=blue, urlcolor=blue]{hyperref}
\usepackage{framed,multirow}
\usepackage{threeparttable}
\usepackage[misc]{ifsym}
\usepackage[symbol]{footmisc}
\usepackage{amsmath}
\usepackage{amssymb}

\begin{document}
\title{Semi-supervised Left Atrium Segmentation with Mutual Consistency Training}
\titlerunning{MC-Net for Semi-supervised Segmentation}

\author{Yicheng Wu\inst{1,2} \and Minfeng Xu \inst{1} \and Zongyuan Ge\inst{3,4} \and Jianfei Cai \textsuperscript{2(\Letter)} \and Lei Zhang\inst{1}}
\authorrunning{Yicheng Wu et al.}
\institute{\textsuperscript{1}DAMO Academy, Alibaba Group, Hangzhou, 311121, China\\
\textsuperscript{2}Department of Data Science \& AI, Faculty of Information Technology, \\Monash University, Melbourne, VIC 3800, Australia\\
\textsuperscript{3}Monash-Airdoc Research, Monash University, Melbourne, VIC 3800, Australia\\
\textsuperscript{4}Monash Medical AI, Monash eResearch Centre, Melbourne, VIC 3800, Australia \\
\email{jianfei.cai@monash.edu}
}
\maketitle 
\begin{abstract}
Semi-supervised learning has attracted great attention in the field of machine learning, especially for medical image segmentation tasks, since it alleviates the heavy burden of collecting abundant densely annotated data for training. However, most of existing methods underestimate the importance of challenging regions (e.g. small branches or blurred edges) during training. We believe that these unlabeled regions may contain more crucial information to minimize the uncertainty prediction for the model and should be emphasized in the training process. Therefore, in this paper, we propose a novel Mutual Consistency Network (MC-Net) for semi-supervised left atrium segmentation from 3D MR images. Particularly, our MC-Net consists of one encoder and two slightly different decoders, and the prediction discrepancies of two decoders are transformed as an unsupervised loss by our designed cycled pseudo label scheme to encourage mutual consistency. Such mutual consistency encourages the two decoders to have consistent and low-entropy predictions and enables the model to gradually capture generalized features from these unlabeled challenging regions. We evaluate our MC-Net on the public Left Atrium (LA) database and it obtains impressive performance gains by exploiting the unlabeled data effectively. Our MC-Net outperforms six recent semi-supervised methods for left atrium segmentation, and sets the new state-of-the-art performance on the LA database.
\end{abstract}

\section{Introduction}
Accurately segmenting the organs or tissues is a fundamental and significant step to construct a computer-aided diagnosis (CAD) system, which plays a critical role for medical image quantitative analysis. Most of existing segmentation methods rely on abundant labeled data for training, where collecting a large number of labeled data is labor-intensive and time-consuming. Considering collecting unlabeled data is much easier, it is highly desirable to develop semi-supervised segmentation methods to effectively exploit rich unlabeled data.

A few recent semi-supervised models \cite{fixmatch,cct} have been proposed to study the consistency regularization. For example, Sohn et al. \cite{fixmatch} applied weak or strong perturbations to augment data and constrained the model to output invariant results over different perturbations. Lee et al. \cite{pseudo} employed pseudo labels to encourage a low-density separation between classes as the entropy regularization for training. These semi-supervised methods have achieved promising progresses.

For medical tasks, there are also several semi-supervised works to segment human organs. For instance, Yu et al. \cite{uamt} designed a teacher-student model to segment left atrium. Enforcing the consistency between the student model and the teacher model can facilitate the model learning. Li et al. \cite{sassnet} introduced an adversarial loss to encourage that the feature spaces of labeled data and unlabeled data are close. Luo et al. \cite{dtc} studied the relation between semantic segmentation and shape regression to leverage unlabeled data. Xie et al. \cite{pairwise} utilized attention mechanisms to calculate the semantic similarities between labeled data and unlabeled data. Fang et al. \cite{dmnet} used a co-training framework to boost each sub-model and also adopted an adversarial loss to further improve the performance. However, these deep models either require extra components to obtain performance gains or underestimate the importance of some challenging regions (e.g. small branches or adhesive edges around the target) in the training process.
\begin{figure*}[htb]
\centering
\includegraphics[width=1\textwidth]{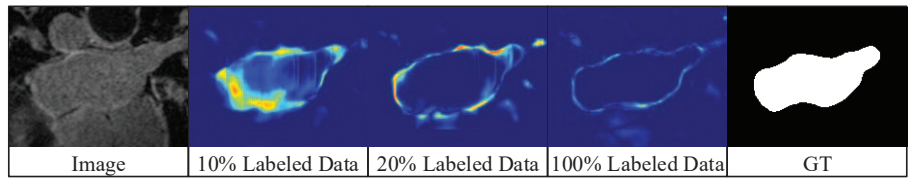}
\caption{\label{uncertainty}Example MR image (1st column), the uncertainty of deep models trained with 10\% labeled data (2nd column), 20\% labeled data (3rd column) and all labeled data (4th column), and corresponding ground truth (5th column) on the LA database.}
\end{figure*}

Specifically, due to the limitation of training data, deep models often have ambiguous predictions in some complex or blurred areas. For example, Figure~\ref{uncertainty} gives the comparisons of uncertainty of deep models trained with 10\% labeled data, 20\% labeled data and all labeled data on the Left Atrium (LA) database \cite{data}. 
First, we observe that \emph{\textbf{(1) these ambiguous predictions mainly locate at challenging areas}}. \cite{uamt} claimed that these ambiguous targets may lead to meaningless and unreliable guidance. However, we believe that these challenging regions contain more crucial information and should be emphasized during training since hard examples can make training more efficient \cite{ohem} and most of other regions can be correctly segmented by deep models even trained with 10\% labeled data.
Second, Figure~\ref{uncertainty} indicates that \emph{\textbf{(2) deep models trained by more labeled data tend to output less ambiguous predictions}}. The ambiguous predictions can be represented by the model-based epistemic uncertainty \cite{uncertainty}. We deem that the epistemic uncertainty is able to evaluate the model's generalization ability since the most generalized model, i.e. the one trained with all labeled data, has the minimum uncertainty (see the fourth column in Figure~\ref{uncertainty}).
Hence, based on the observations \emph{\textbf{(1)}} and \emph{\textbf{(2)}}, we hypothesize that such uncertainty information can be regarded as additional supervision signals to boost model training. In other words, these most challenging and valuable regions should be explored to train the segmentation model even without labels.
 
Therefore, in this paper, we propose a novel Mutual Consistency model (MC-Net, see Figure~\ref{MCNet}) for semi-supervised left atrium segmentation from 3D MR images. Our MC-Net is composed of one encoder and two slightly different decoders and the discrepancy of two outputs is used to capture the uncertainty information.
Then, we employ a sharpening function \cite{fixmatch} to generate soft pseudo labels from the probability outputs as the entropy regularization \cite{pseudo}.
Afterwards, we design a new cycled pseudo label scheme to
leverage the uncertainty information to learn more critical features by encouraging mutual consistency \cite{mutual}.
Such mutual consistency constrains MC-Net to generate consistent and low-entropy results of the same input so that the model-based epistemic uncertainty is reduced, which enables the model to learn generalized feature representation from these unlabeled challenging regions \cite{cct,fixmatch}.
We evaluate our proposed MC-Net against six recent state-of-the-art (SOTA) methods on the popular LA database \cite{data}. The experiment results reveal that the designed MC-Net outperforms all other existing semi-supervised segmentation methods. The ablation studies further demonstrate the effectiveness of each component design.

The contributions of our model include: (1) we explore the model-based uncertainty information to emphasize the unlabeled challenging regions during training; (2) we design a novel cycled pseudo label scheme to facilitate the model training by encouraging mutual consistency; (3) to our best knowledge, the proposed MC-Net has achieved the new state-of-the-art performance in the semi-supervised left atrium segmentation task on the LA database.

\section{Method}
\begin{figure*}[htb]
\centering
\includegraphics[width=1\textwidth]{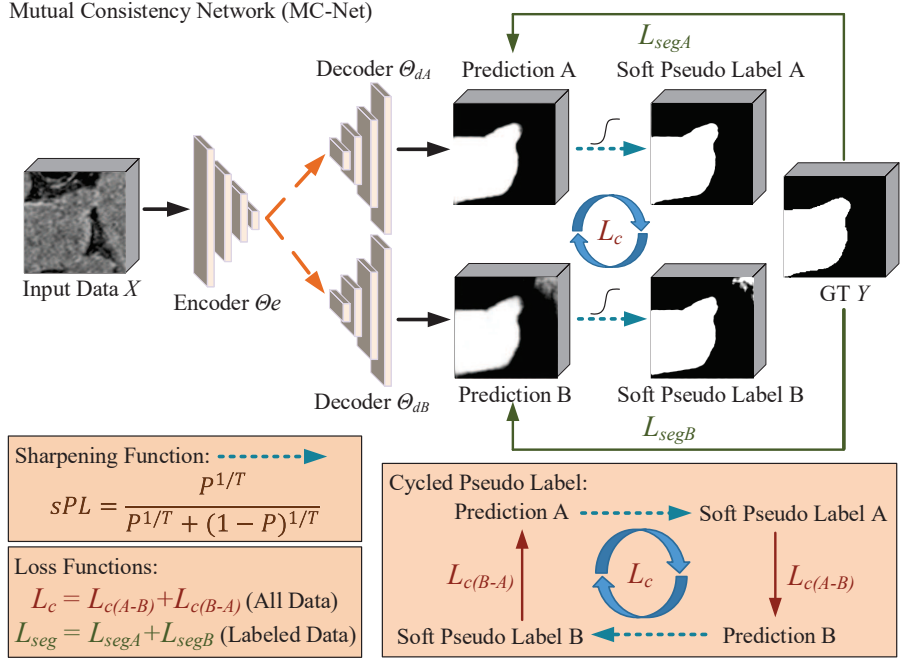}
\caption{\label{MCNet}Diagram of our proposed MC-Net.}
\end{figure*}

As shown in Figure~\ref{MCNet}, our MC-Net has two unique attributes. First, we embed an extra slightly different decoder into the original V-Net \cite{vnet} and utilize the discrepancy of the two decoder outputs to capture the uncertainty information. Second, we design the cycled pseudo label scheme to transform such uncertainty as an unsupervised loss to encourage mutual consistency for training.
\subsection{Model Architecture}
There are several representative methods to measure the epistemic uncertainty. For example, the Monte Carlo dropout \cite{uncertainty} is a popular one. Given a 3D input sample $X \in \mathbb{R}^{H\times W\times D}$, we can perform $N$ stochastic forward passes with random dropout, where the dropout layer is able to sample sub-models $\theta_{n}$ from the original model $\theta$. In this way, the deep model $\theta$ outputs a set of probability vector: $\{P_{n}\}_{n=1}^{N}$. The uncertainty $u$ can be approximated by the statistics of the predictions of all sub-models $\theta_{n}$. For instance, a related work \cite{uamt} employs the Monte Carlo dropout to estimate the uncertainty $u$ as:
\begin{equation}
\mu_{c} = \frac{1}{N}\sum_{n} P_{n}^{c}, \quad
u=-\sum_{C} \mu_{c} log\mu_{c}
\end{equation}
where $P_{n}^{c}$ represents the output of the $c$-th class in the $n$-th time, $\mu_{c}$ is the mean of $N$ predictions, $C$ is the number of classes and the uncertainty $u \in \mathbb{R}^{H\times W\times D}$ is essentially the voxel-wise entropy. One of the issues of such a method is that it requires multiple inferences, e.g. eight stochastic forward passes every iteration in \cite{uamt}, to estimate the uncertainty, which brings more computational costs.

Therefore, inspired by \cite{dass}, we simplify the Monte Carlo dropout via introducing an auxiliary classifier, which reduces both training and inference costs. Specifically, we design two slightly different decoders to approximate the epistemic uncertainty. One decoder $\theta_{dA}$ employs original transposed convolution for up-sampling like V-Net and another decoder $\theta_{dB}$ uses tri-linear interpolation to expand the feature maps as an auxiliary classifier. Other modules are identical to V-Net. 
The two decoders receive the same deep features $F_{e}$ from encoder $\theta_{e}$ and then generate two features $F_{A}$ and $F_{B}$. This process can be formulated as
\begin{equation}
F_{e} = f_{\theta_{e}}(X), \quad
F_{A} = f_{\theta_{dA}}(F_{e}), \quad 
F_{B} = f_{\theta_{dB}}(F_{e})
\end{equation}
where probability outputs $P_{A}$ and $P_{B}$ can be obtained from deep features $F_{A}$ and $F_{B}$, respectively, with a Sigmoid activation function. Such two slightly different decoders can increase the diversity of segmentation models, and the diversified features in different sub-models are able to reduce over-fitting and improve the performance \cite{cotraining}. Meanwhile, for fair comparisons, we do not further introduce any other additional designs to enhance the V-Net backbone \cite{vnet}.

Compared to Monte Carlo dropout \cite{uncertainty,uamt}, the sub-models $\theta_{n}$ of our MC-Net are fixed and do not require additional perturbations in the training process. As shown in Figure~\ref{MCNet}, the model-based epistemic uncertainty is approximated by the discrepancy between the two decoder outputs $P_{A}$ and $P_{B}$.

\subsection{Cycled Pseudo Label}
Based on such model design, we then transform the prediction discrepancies as auxiliary supervision signals to boost training. First, we use a sharpening function \cite{sharpening} to transform probability outputs $P_{A}$ and $P_{B}$ into soft pseudo labels $sPL_{A}$ and $sPL_{B}$ $\in [0,1]^{H\times W\times D}$. The sharpening function is defined as 
\begin{equation}
sPL = \frac{P^{1/T}}{P^{1/T}+(1-P)^{1/T}}
\end{equation}
where $T$ is a constant to control the temperature of sharpening. The soft pseudo labels make contribution to entropy regularization for training \cite{pseudo}. Compared with pseudo labels generating by a fixed threshold, soft pseudo labels are able to eliminate the impacts of some mislabeled training data \cite{sharpening}.

Afterwards, we employ $sPL_{A}$ to supervise $P_{B}$ and then employ $sPL_{B}$ to supervise $P_{A}$ to achieve mutual consistency \cite{mutual}. In this way, the two decoders can learn from each other and be trained in an `end-to-end' manner. Via this design, the predictions $P_{A}$ and $P_{B}$ are encouraged to be consistent and of low entropy. Such consistency and entropy regularization facilitate the model to pay more attention to the unlabeled challenging and uncertain areas \cite{cct,fixmatch}.

Hence, our MC-Net can be trained via the weighted sum of a segmentation loss $L_{seg}$ and a consistency loss $L_c$. The total loss can be written as
\begin{equation}
\begin{matrix} 
loss =  \underbrace{Dice(P_{A},Y) + Dice(P_{B},Y)}_{{}L_{seg}} \end{matrix} \begin{matrix}+\lambda\times \underbrace{(L_2(P_{A}, sPL_{B})+L_2(P_{B}, sPL_{A})) }_{{}L_{c}}\end{matrix}
\end{equation}
where $Dice$ represents the $Dice$ loss, $L_2$ is the Mean Squared Error (MSE) loss, $Y$ is the ground truth and $\lambda$ is a weight to balance $L_{seg}$ and $L_c$. Note that, $L_{seg}$ is only calculated from labeled data and $L_c$ is unsupervised, which is used to supervise all training data.

\section{Experiment and Results}
\subsection{Database}
We evaluate the proposed MC-Net on the LA database \cite{data} from the 2018 Atrial Segmentation Challenge\footnote{http://atriaseg2018.cardiacatlas.org}. The LA database consists of 100 gadolinium-enhanced MR imaging scans, with a fixed split of 80 samples for training and 20 samples for validation. The isotropic resolution is $0.625\times0.625\times0.625$ mm. We report the performance on the validation set for fair comparisons as \cite{lgermt,sassnet,dtc,duwm,uamt}.
\subsection{Implementation Details}
For pre-processing, we first cropped the 3D MR images with enlarged margins according to the targets. These scans were further normalized as zero mean and unit variance. For training, we randomly cropped 3D patches of size $112\times112\times80$ and applied the 2D rotation and flip operations as the data augmentation. Then, we set the batch size as 4 and each batch included two labeled patches and two unlabeled patches. The temperature constant $T$ was set as 0.1 and the weight $\lambda$ was set as a time-dependent Gaussian warming-up function \cite{rampup}. The proposed MC-Net was trained by a SGD optimizer for 6K iterations, with an initial learning rate 0.01 decayed by 10\% every 2.5K iterations. Most of the experiment settings were set same as the recent SOTA methods \cite{sassnet,dtc,uamt}. 

For testing, we employed a sliding window of size $112\times112\times80$ with a fixed stride ($18\times18\times4$) to extract patches. Then we recomposed the patch-based predictions as entire results. Note that, we used the mean of $P_{A}$ and $P_{B}$ as the final output during testing. All experiments were conducted on the same environments with fixed random seeds (Hardware: Intel Xeon E5-2682 CPU, NVIDIA Tesla P100 GPU; Software: Pytorch 1.7.0+cu110, and Python 3.8.5).
\subsection{Results}
\begin{figure*}[!htb]
\centering
\includegraphics[width=1\textwidth]{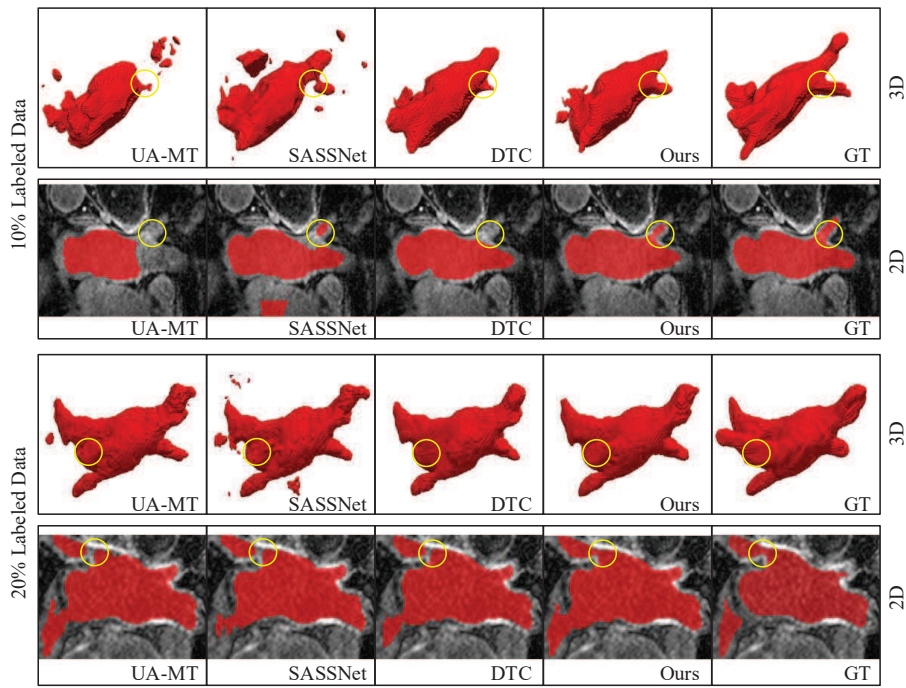}
\caption{\label{Result}Two segmentation results obtained by UA-MT \cite{uamt} (1st column), SASSNet \cite{sassnet} (2nd column), DTC \cite{dtc} (3rd column), our proposed MC-Net (4th column), and the corresponding ground truth (5th column) on the LA database. The comparisons using 10\% or 20\% labeled data are showed on the top or bottom two rows, respectively.}
\end{figure*}
{
\begin{table*}[!htb]
	\centering
	\caption{Comparisons with six state-of-the-art methods on the LA database.}
	\label{tab:tab1}
    \begin{threeparttable}
	\resizebox{\textwidth}{!}{
	\begin{tabular}{c|cc|cccc}
		\hline 
		\hline
		\multirow{2}{*}{Method}&\multicolumn{2}{c}{\# Scans used}&\multicolumn{4}{|c}{Metrics}\\
		\cline{2-7}
		&Labeled&Unlabeled &Dice(\%)&Jaccard(\%)&95HD(voxel)&ASD(voxel)\\
		\hline
		V-Net & 8(10\%) &0 &79.99 &68.12 &21.11 &5.48 \\
		V-Net &16(20\%) &0 &86.03 &76.06 &14.26 &3.51 \\
		V-Net &80(All) &0 &91.14 &83.82 &5.75 &1.52 \\
		\hline
		DAP \cite{dap} (MICCAI'19) & 8(10\%) &72 &81.89 &71.23 &15.81 &3.80 \\
		UA-MT \cite{uamt} (MICCAI'19) & 8(10\%) &72 &84.25 &73.48 &13.84 &3.36 \\
		SASSNet \cite{sassnet} (MICCAI'20) & 8(10\%) &72 &87.32 &77.72 &9.62 &2.55 \\
		LG-ER-MT \cite{lgermt} (MICCAI'20) & 8(10\%) &72 &85.54 &75.12 &13.29 &3.77 \\
		DUWM \cite{duwm} (MICCAI'20) & 8(10\%) &72 &85.91 &75.75 &12.67 &3.31 \\
		DTC \cite{dtc} (AAAI'21)\tnote{*} & 8(10\%) &72 &86.57 &76.55 &14.47 &3.74 \\
		\textbf{MC-Net (Ours)}
		&8(10\%)&72&\textbf{87.71}&\textbf{78.31} &\textbf{9.36} & \textbf{2.18}\\
		\hline
		DAP \cite{dap} (MICCAI'19) &16(20\%) &64 &87.89 &78.72 &9.29 &2.74 \\
		UA-MT \cite{uamt} (MICCAI'19) &16(20\%) &64 &88.88 &80.21 &7.32 &2.26 \\
		SASSNet \cite{sassnet} (MICCAI'20) &16(20\%) &64 &89.54 &81.24 &8.24 &2.20\\
		LG-ER-MT \cite{lgermt} (MICCAI'20) &16(20\%) &64 &89.62 &81.31 &7.16 &2.06 \\
		DUWM \cite{duwm} (MICCAI'20) &16(20\%) &64 &89.65 &81.35 &7.04 &2.03 \\
		DTC \cite{dtc} (AAAI'21) &16(20\%) &64 &89.42 &80.98 &7.32 &2.10 \\
		\textbf{MC-Net (Ours)}
		&16(20\%)&64&\textbf{90.34}&\textbf{82.48} &\textbf{6.00} & \textbf{1.77}\\
		\hline
		\hline
	\end{tabular}}
    \begin{tablenotes}
    \footnotesize
    \item[*]Since the results of DTC model using 10\% labeled data with 90\% unlabeled data\\ were not given, we conducted the experiments on the LA database as \cite{dtc}.
    \end{tablenotes}
    \end{threeparttable}
\end{table*}
}
Figure~\ref{Result} shows the results obtained by UA-MT \cite{uamt}, SASSNet \cite{sassnet}, DTC \cite{dtc}, our MC-Net, and the corresponding ground truth on the LA database from left to right. There are two popular semi-supervised settings that the first uses 10\% labeled data with 90\% unlabeled data and the second uses 20\% labeled data with 80\% unlabeled data. The corresponding results are also shown in Figure~\ref{Result}. It can be seen that our model generates more complete left atrium than all other existing SOTA methods in either 3D or 2D view. Note that, we do not use any morphology algorithms as the post-processing module to refine the results. The MC-Net naturally generates better results in some challenging areas (see yellow circles) and eliminates most of isolated regions on the LA database. Such ability is essential for further medical quantitative analysis.

Table~\ref{tab:tab1} shows the quantitative results in terms of the Dice, Jaccard, 95\% Hausdorff Distance (95HD) and average surface distance (ASD). It also gives the results of V-Net under fully supervised settings (with 10\%, 20\% and all labeled data) as the reference. By exploiting unlabeled data effectively, our proposed MC-Net produces impressive performance gains and our method with only 20\% labeled training data obtain comparable results, e.g. 90.34\% vs. 91.14\% of Dice, with the upper bound (V-Net with 100\% labeled training data). It can also be seen from Table~\ref{tab:tab1} that our MC-Net significantly outperforms the six recent SOTA methods on the LA database in all semi-supervised settings.

\subsection{Ablation Study}
The ablation studies (see Table~\ref{tab:tab2}) were conducted to show the effectiveness of each component on the LA database. We can see from Table~\ref{tab:tab2} that, in each semi-supervised setting, the one with two slightly different decoders (V2d-Net) generates better results than the one with two identical decoders (V2-Net). This suggests that increasing the diversity of the segmentation model can improve the performance. Moreover, encouraging consistent results, i.e. minimizing the differences between $sPL_{A}$ and $sPL_{B}$ (labeled as +sPL), in general leads to the performance gains and applying the entropy regularization, i.e. using the cycled pseudo label (labeled as +CPL), is able to further improve the performance. Note that, our MC-Net can be easily combined with other shape-constrained models \cite{sassnet,dtc} to enhance the segmentation results.
{
\begin{table*}[!htb]
	\centering
	\caption{Ablation studies of our proposed MC-Net on the LA database.}
	\label{tab:tab2}
    \begin{threeparttable}
	\resizebox{\textwidth}{!}{
	\begin{tabular}{c|cc|cccc}
		\hline 
		\hline
		\multirow{2}{*}{Method}&\multicolumn{2}{c}{\# Scans used}&\multicolumn{4}{|c}{Metrics}\\
		\cline{2-7}
		&Labeled&Unlabeled &Dice(\%)&Jaccard(\%)&95HD(voxel)&ASD(voxel)\\
		\hline
		V2-Net  & 8(10\%) &72 &84.95 &74.27 &15.21 &4.56 \\
		V2d-Net & 8(10\%) &72 &85.79 &75.41 &14.45 &3.83  \\
		V2-Net+sPL  & 8(10\%) &72 &86.61 &76.65 &13.39 &3.93 \\
		V2d-Net+sPL & 8(10\%) &72 &86.84 &76.97 &14.16 &4.05  \\
		V2-Net+CPL & 8(10\%) &72 &86.69 &76.74 &12.66 &3.20  \\
		V2d-Net+CPL
		&8(10\%)&72&\textbf{87.71}&\textbf{78.31} &\textbf{9.36} & \textbf{2.18}\\
		\hline
		V2-Net  & 16(20\%) &64 &88.14 &79.10 &14.22 &3.78 \\
	    V2d-Net & 16(20\%) &64 &88.98 &80.36 &7.61 &2.25  \\
		V2-Net+sPL  & 16(20\%) &64 &88.55 &79.64 &14.06 &3.58 \\
		V2d-Net+sPL & 16(20\%) &64 &90.15 &82.15 &6.20 &1.89  \\
		V2-Net+CPL & 16(20\%) &64 &89.66 &81.36 &11.14 &2.86  \\
		V2d-Net+CPL
		&16(20\%)&64&\textbf{90.34}&\textbf{82.48} &\textbf{6.00} & \textbf{1.77}\\
		\hline
		\hline
	\end{tabular}}
    \end{threeparttable}
\end{table*}
}
\section{Conclusion}
In this paper, we have presented a mutual consistency network (MC-Net) for semi-supervised left atrium segmentation. Our key idea is that the unlabeled challenging regions should play a critical role in semi-supervised learning. Hence, via the designed cycled pseudo label scheme, our model is encouraged to generate consistent and low-entropy predictions so that more generalized features can be captured from these crucial areas to boost the model training.
The proposed MC-Net achieves, to our best knowledge, the most accurate semi-supervised left atrium segmentation performance on the LA database.
\subsubsection{Acknowledgements:}
This work was done during an internship at Alibaba Group and was partially supported by Monash FIT Start-up Grant. We also appreciate the efforts devoted to collect and share the LA database \cite{data} and several available repositories \cite{sassnet,dtc,uamt}.

\bibliographystyle{splncs04}
\bibliography{paper.bib}
\newpage
\section{Supplementary Material}
{
\begin{table*}[!htb]
	\centering
	\caption{Comparisons of the two decoder outputs of our proposed MC-Net on LA. Note that, the model only with the original decoder $\theta_{dA}$ also achieves the superior results without extra computational complexities on LA.}
	\label{tab3}
    \begin{threeparttable}
	\resizebox{\textwidth}{!}{
	\begin{tabular}{c|cc|cccc|cc}
		\hline 
		\hline
		\multirow{2}{*}{Method}&\multicolumn{2}{c}{\# Scans used}&\multicolumn{4}{|c}{Metrics}&\multicolumn{2}{|c}{Complexity}\\
		\cline{2-9}
		&Labeled&Unlabeled &Dice(\%)&Jaccard(\%)&95HD(voxel)&ASD(voxel)&Para.&MACs\\
		\hline
		V-Net & 8(10\%) &0 &79.99 &68.12 &21.11 &5.48 &9.45M&47.17G \\
		Decoder $\theta_{dA}$ & 8(10\%) &72 &87.43 &77.87 &11.18 &2.27 &9.45M&47.17G \\
		Decoder $\theta_{dB}$ & 8(10\%) &72 &\textbf{87.83} &\textbf{78.49} &\textbf{9.07} &2.27 &10.28M&65.51G\\
		MC-Net &8(10\%)&72&87.71&78.31 &9.36 & \textbf{2.18} &12.35M&95.11G\\
		\hline
		V-Net &16(20\%) &0 &86.03 &76.06 &14.26 &3.51 &9.45M&47.17G\\
		Decoder $\theta_{dA}$ & 16(20\%) &64 &90.25 &82.32 &6.16 &1.95  &9.45M&47.17G\\
		Decoder $\theta_{dB}$ & 16(20\%) &64 &90.27 &82.37 &6.04 &1.72 &10.28M&65.51G\\
		MC-Net &16(20\%)&64&\textbf{90.34}&\textbf{82.48} &\textbf{6.00} & \textbf{1.77} &12.35M&95.11G\\
		\hline
		V-Net &80(All) &0 &91.14 &83.82 &5.75 &1.52  &9.45M&47.17G\\
		\hline
		\hline
	\end{tabular}}
    \end{threeparttable}
\end{table*}
}
{
\begin{table*}[!htb]
	\centering
	\caption{Comparisons with three SOTA methods on the Pancreas-CT database.}
	\label{tab4}
    \begin{threeparttable}
	\resizebox{\textwidth}{!}{
	\begin{tabular}{c|cc|cccc}
		\hline 
		\hline
		\multirow{2}{*}{Method}&\multicolumn{2}{c}{\# Scans used}&\multicolumn{4}{|c}{Metrics}\\
		\cline{2-7}
		&Labeled&Unlabeled &Dice(\%)$\uparrow$ &Jaccard(\%)$\uparrow$&95HD(voxel)$\downarrow$&ASD(voxel)$\downarrow$\\
		\hline
		V-Net\tnote{*} &6 (10\%) &0 &55.06 &40.48 &32.80 &12.67 \\
		V-Net\tnote{*} &12 (20\%) &0 &69.65 &55.19 &20.20 &6.31 \\
		V-Net\tnote{*} &62 (All) &0 &83.02 &71.36 &5.18 &1.19 \\
		\hline
		UA-MT\tnote{*} (MICCAI'19)& 6 (10\%) &56 (90\%) &68.70 &54.65 &\textbf{13.89} &3.23 \\
		SASSNet\tnote{*} (MICCAI'20)  &6 (10\%)  &56 (90\%) &66.52 &52.23 &17.11 &\textbf{2.25}\\
		DTC\tnote{*} (AAAI'21)  &6 (10\%)  &56 (90\%) &66.27 &52.07 &15.00 &4.44 \\
		MC-Net\tnote{*} (2021)  &6 (10\%)  &56 (90\%) &\textbf{68.94} &\textbf{54.74} &16.28 &3.16\\
		\hline
		UA-MT\tnote{*} (MICCAI'19) & 12 (20\%) &50 (80\%) &76.75 &63.77 &11.41 &2.79 \\
		SASSNet\tnote{*} (MICCAI'20)  &12 (20\%)  &50 (80\%) &77.11 &64.24 &8.93 &\textbf{1.41}\\
		DTC\tnote{*} (AAAI'21)  &12 (20\%)  &50 (80\%) &78.27 &64.75 &\textbf{8.36} &2.25 \\
		MC-Net\tnote{*} (2021)  &12 (20\%)  &50 (80\%) &\textbf{79.05}&\textbf{65.83}&10.29&2.72\\
		\hline
		\hline
	\end{tabular}}
    \begin{tablenotes}
    \footnotesize
    \item[*]All experiments were conducted on the same environments.\\
    Pancreas-CT: https://wiki.cancerimagingarchive.net/display/Public/Pancreas-CT
    \end{tablenotes}
    \end{threeparttable}
\end{table*}
}
\begin{figure*}[htb]
\centering
\includegraphics[width=1\textwidth]{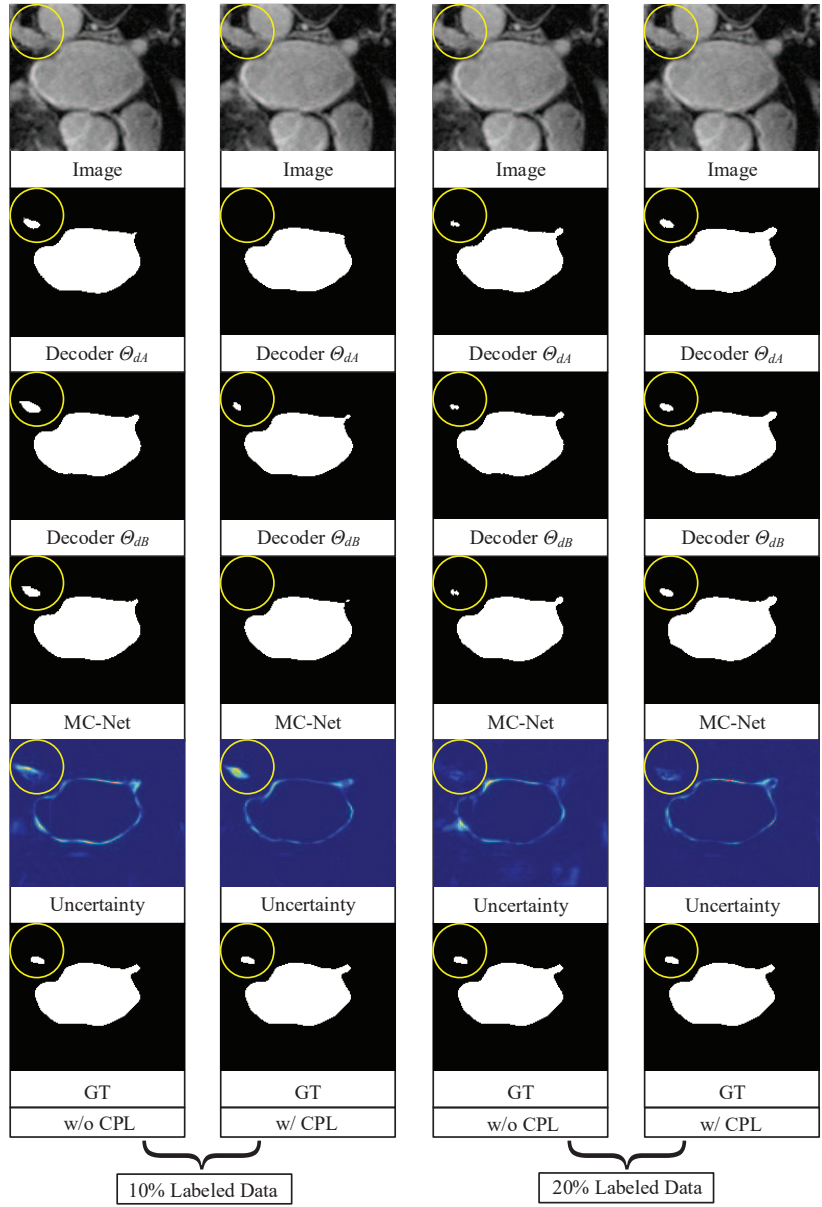}
\caption{\label{Uncertainty2}Comparisons of the two decoder outputs of our proposed MC-Net on the LA database. There are the example MR images (1st row), segmentation results of decoder $\theta_{dA}$ (2nd row) and decoder $\theta_{dB}$ (3rd row), mean results of $\theta_{dA}$ and $\theta_{dB}$ i.e. our MC-Net, the corresponding estimated uncertainty and ground truth. Note that, we also show the segmentation results of the MC-Net with and without the designed cycled pseudo label scheme (labeled as w/ CPL and w/o CPL, respectively).}
\end{figure*}
\end{document}